\pdfoutput=1

\documentclass[11pt]{article}

\usepackage[preprint]{acl}

\usepackage{times}
\usepackage{latexsym}
\usepackage{tcolorbox}
\usepackage{multicol}
\usepackage{booktabs}
\usepackage{natbib}
\usepackage{float}
\usepackage{listings}
\usepackage{xcolor}

\usepackage[T1]{fontenc}

\usepackage[utf8]{inputenc}

\usepackage{microtype}

\usepackage{inconsolata}

\usepackage{graphicx}

\lstdefinelanguage{json}{
    basicstyle=\ttfamily\small,
    numbers=left,
    numberstyle=\tiny\color{gray},
    stepnumber=1,
    numbersep=5pt,
    showstringspaces=false,
    breaklines=true,
    frame=lines,
    backgroundcolor=\color{white},
    keywordstyle=\color{blue},
    stringstyle=\color{orange},
    morestring=[b]",
}

\title{Deconstructing Long Chain-of-Thought: A Structured Reasoning Optimization Framework for Long CoT Distillation}

\author{
 \textbf{Yijia Luo\textsuperscript{1*}},
 \textbf{Yulin Song\textsuperscript{1,2*}},
 \textbf{Xingyao Zhang\textsuperscript{1*}},
 \textbf{Jiaheng Liu\textsuperscript{1†}},
\\
 \textbf{Weixun Wang\textsuperscript{1}},
 \textbf{GengRu Chen\textsuperscript{1}},
 \textbf{Wenbo Su\textsuperscript{1}},
 \textbf{Bo Zheng \textsuperscript{1}}
\\
\\
 \textsuperscript{1}Alibaba Group,
 \textsuperscript{2}New York University
\\
 \small{
   \textbf{Correspondence:} \href{mailto:luoyijia.lyj@taobao.com}{\{luoyijia.lyj, songyulin.syl, ljh411989\}@alibaba-inc.com}
 }
}

\begin{document}
\maketitle

\let\thefootnote\relax\footnotetext{* First three authors contributed equally. }
\let\thefootnote\relax\footnotetext{† Corresponding Author: Jiaheng Liu.}

\begin{abstract}
Recent advancements in large language models (LLMs) have demonstrated remarkable reasoning capabilities through long chain-of-thought (CoT) reasoning. The R1 distillation scheme has emerged as a promising approach for training cost-effective models with enhanced reasoning abilities. However, the underlying mechanisms driving its effectiveness remain unclear. This study examines the universality of distillation data and identifies key components that enable the efficient transfer of long-chain reasoning capabilities in LLM distillation. Our findings reveal that the effectiveness of long CoT reasoning distillation from teacher models like QwQ degrades significantly on nonhomologous models, challenging the assumed universality of current distillation methods. To gain deeper insights into the structure and patterns of long CoT reasoning, we propose \textbf{DLCoT (Deconstructing Long Chain-of-Thought)}—a distillation data enhancement framework. DLCoT consists of three key steps: (1) \textbf{data segmentation} to decompose complex long CoT structures, (2) \textbf{simplification} by eliminating unsolvable and redundant solutions, and (3) \textbf{optimization} of intermediate error states. Our approach significantly improves model performance and token efficiency, facilitating the development of high-performance LLMs. Code and data will be released at: https://github.com/elena-luo/SODE.git.
\end{abstract}

\section{Introduction}
\begin{figure}[t]
  \centering
  \includegraphics[width=\columnwidth]{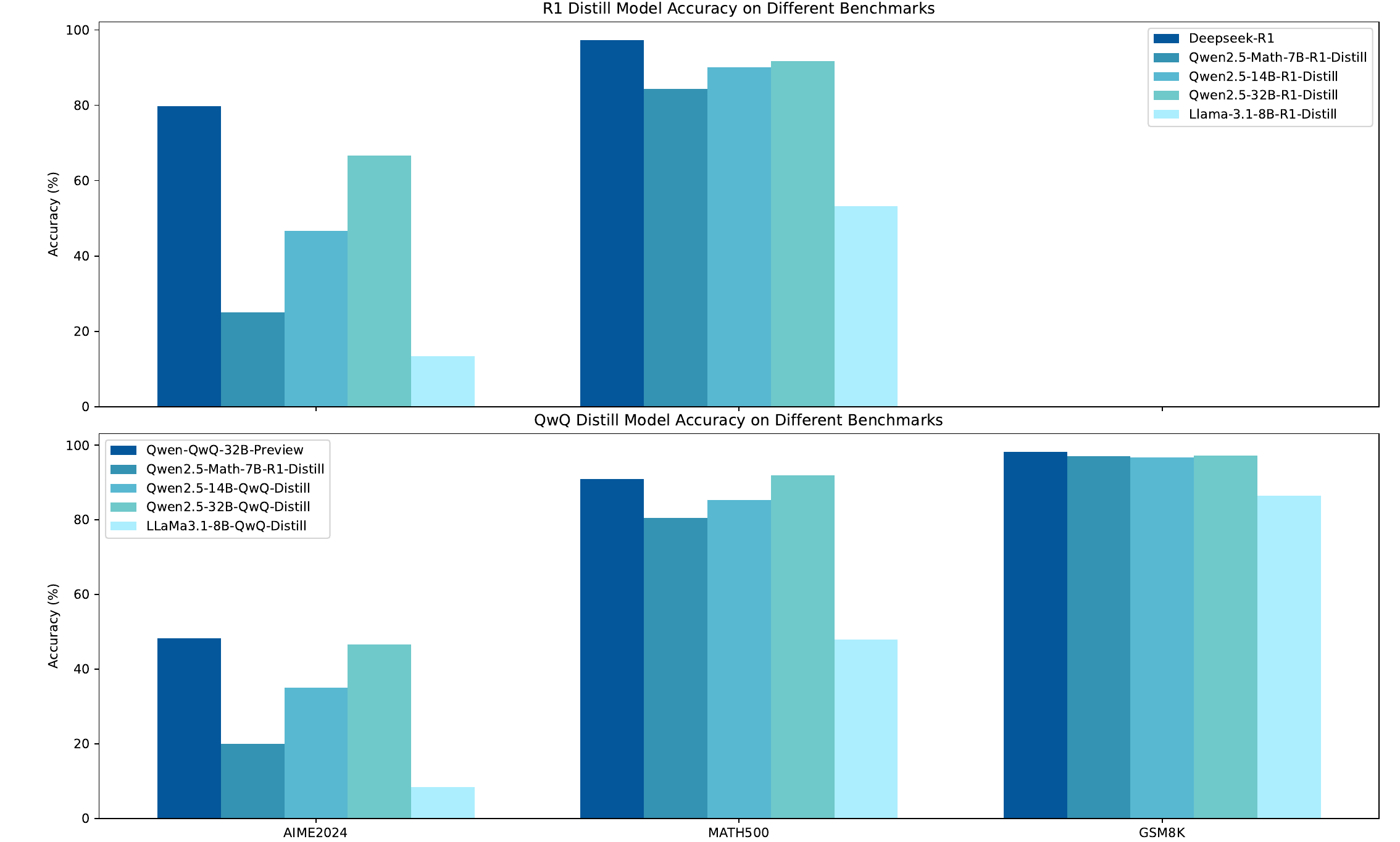}
  \caption{Comparative Analysis of Model Distillation Performance Using 6K NuminaMath Correct Answer data.
Upper Panel: Qwen and Llama distilled through R1-generated data. Notably, while these fail to fully replicate the ability R1 reported, the Qwen2.5-14B achieves comparable accuracy to the QwQ-distilled Qwen2.5-32B, demonstrating R1's enhanced cross-model transferability in distillation scenarios.
Lower Panel: The Qwen-family models demonstrate superior distillation efficacy compared to Llama when using data generated by QwQ. The Qwen2.5-32B achieves performance parity with the QwQ.}
  \label{fig:results}
\end{figure}
Large language models (LLMs) have revolutionized natural language processing (NLP) \cite{openai2024gpt4technicalreport}. A key advancement is the development of chain-of-thought (CoT) prompting \cite{wei2023chainofthoughtpromptingelicitsreasoning, nye2021workscratchpadsintermediatecomputation}, which significantly enhances LLMs' reasoning abilities. Models like O1 \cite{o1report}, R1 \cite{deepseekai2025deepseekr1incentivizingreasoningcapability}, and QwQ \cite{qwq} leverage CoT prompting to excel at complex, multi-step reasoning tasks, featuring with extremely long CoT.

\noindent Despite the remarkable capabilities of long CoT reasoning, training language models to exhibit such reasoning demands substantial computational resources. This raises a critical research question: how can we train these models more efficiently and cost-effectively? For closed-source models like OpenAI’s O1 and Google’s Gemini 2.0 Flash, direct training remains inaccessible to developers. Meanwhile, open-source alternatives like DeepSeek’s R1 show significant challenges due to their massive parameter sizes and the complexity of reinforcement learning, making replication impractical for most practitioners.

\noindent The recently proposed R1 model from Deepseek \cite{deepseekai2025deepseekr1incentivizingreasoningcapability} demonstrates that direct distillation can also achieve superior reasoning performance. Distillation \cite{hinton2015distillingknowledgeneuralnetwork}, a technique for transferring knowledge from a large teacher model to a smaller student model, offers a promising avenue for improving efficiency. However, despite these advancements, the fundamental mechanisms by which long CoT data leads to improved reasoning over traditional, shorter CoTs remain poorly understood.

\noindent We investigate this question by conducting a systematic distillation study across open-source models of varying scales and architectures, focusing particularly on the performance of R1 and QwQ (Figure~\ref{fig:results}). Our findings reveal that the R1-distilled \texttt{Qwen2.5-14B} model \cite{qwen2025qwen25technicalreport} achieves comparable accuracy to the QwQ-distilled \texttt{Qwen2.5-32B} model. This suggests that distillation of R1-based long CoTs offers superior training efficiency and knowledge transferability compared to the QwQ ones.

\begin{figure*}[!h]
  \centering
  \includegraphics[width=\textwidth, trim=0.5cm 3.5cm 0.5cm 1cm, clip]{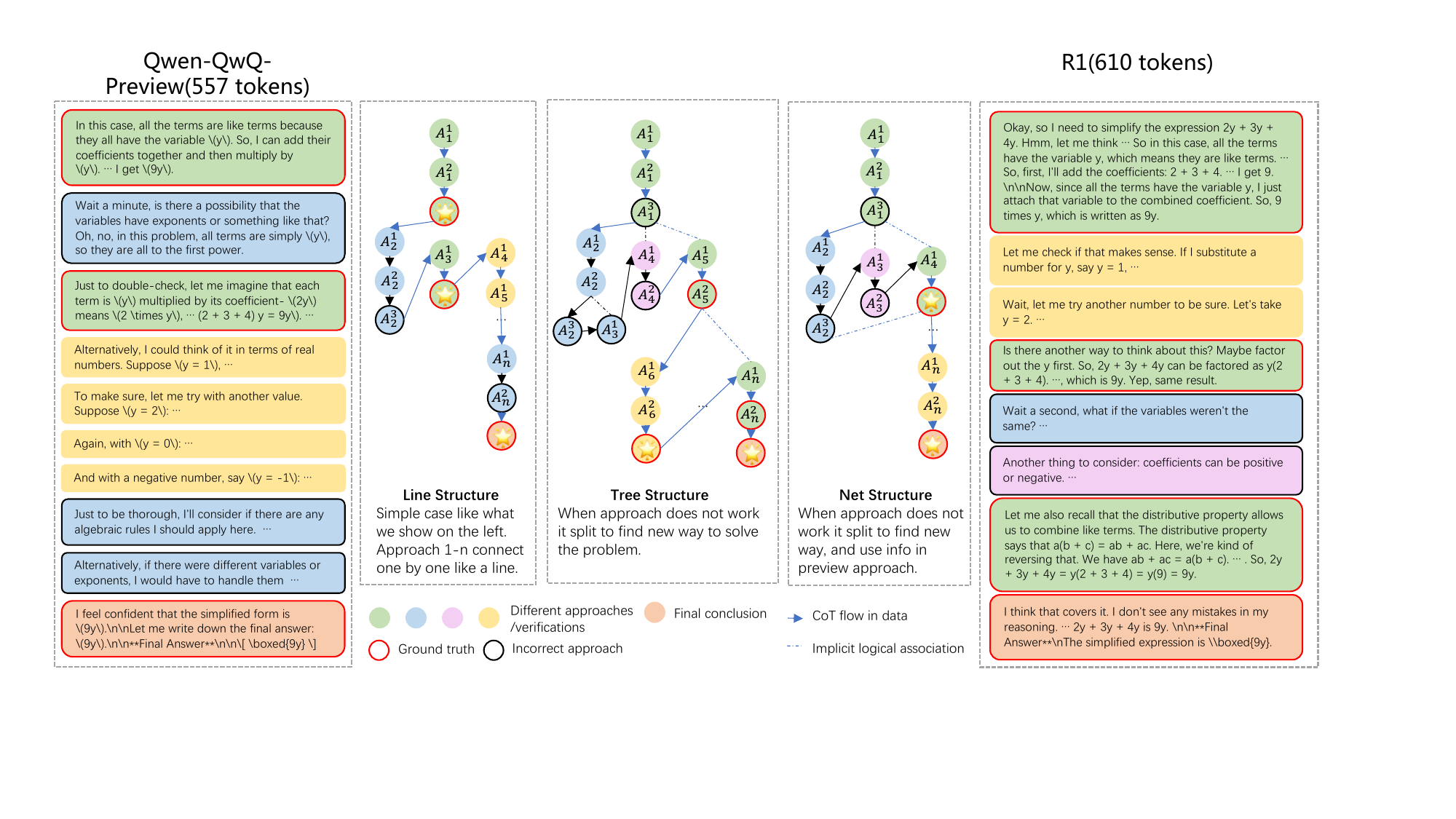}
  \caption{Figure shows an example of QwQ and R1 on "Simplify $2y + 3y + 4y$". Different color-coded text blocks represent distinct solution/verification types. In the middle are three special structure we figure out from long CoT data.
}
  \label{fig:overthinking}
\end{figure*}

\noindent To understand this efficiency gap, we focus on the analysis of the structure of the "thinking" part of R1 and QwQ. Unlike traditional short CoTs, which typically explore a single reasoning path, long CoTs often explore multiple approaches and verifications, exhibiting structural patterns, namely as linear, tree, and network structures (Figure~\ref{fig:overthinking}). These structures are all built upon repeating schemes and a core "trunk", where the trunk represents the shortest complete reasoning chain leading to the correct answer. For instance, in a complex network structure with a complete CoT represented as $A_1A_2...A_n$, the correct answer might be derivable from a shorter sequence like $A_1+A_4$; we define $A_1+A_4$ as the trunk. Our investigation focuses on determining the optimal CoT length—specifically, whether the complete CoT or the trunk represents the "key logic". Furthermore, we observe a prevalent "overthinking" phenomenon \cite{chen2025think23overthinkingo1like} in state-of-the-art LLMs, including Deepseek's R1 and QwQ (Figure~\ref{fig:overthinking}). Student models trained on these overthinking teacher models inherit this characteristic, leading to inefficient resource utilization on simpler tasks. More critically, this inefficient and repetitive exploration of solutions can induce performance degradation in smaller models, often manifesting as repetitive generation patterns.

\noindent To address these challenges, we propose DLCoT (Deconstructing Long Chain-of-Thought), an automated data processing framework specifically designed for long CoT distillation. Guided by our analysis, DLCoT integrates three core modules: Intelligent Segmentation, Redundancy Elimination, and Error Correction. Our experiments demonstrate that prioritizing the reasoning trunk while eliminating redundant paths significantly improves distillation efficiency. Interestingly, removing unique, incorrect reasoning paths or detailed calculation/derivation errors does not provide additional benefits. This suggests that the structural richness and diversity of explored approaches are more valuable for distillation than the precise details of individual reasoning steps.

\noindent Our principal contributions are threefold:
\begin{itemize}
\item \textbf{Distillation Validation}: We systematically explore the feasibility of distilling knowledge from long CoTs using small-scale datasets, evaluating across various source models.
\item \textbf{DLCoT Method}: We develop an automated long
CoT data processing framework, achieving at least $5\%$ improvement in token efficiency while improve accuracy of baseline performance across all benchmarks(see Section ~\ref{sec:Distillation2}).
\item \textbf{Key insights on long CoT}: We identify the key to effective long CoT distillation: the richness of approaches within the reasoning trunk. A diverse set of explored approaches is crucial for stimulating and enhancing the model's reasoning capabilities.
\end{itemize}

\section{Related Work}
\subsection{Test-Time Scaling}
It has been proven effective that CoT enhances the mathematical reasoning of capabilities of LLMs \cite{wei2023chainofthoughtpromptingelicitsreasoning}. Rather than focusing on scaling training-time compute, recent research has emphasized scaling test-time compute, specifically by training reasoning models to generate longer reasoning chains to achieve improved performance. Existing work in this area can be broadly categorized into two main approaches: 1) navigating the model to explore the search space more thoroughly, and 2) adopting human-like reasoning strategies like planning, self-reflection, and self-verification, to train models to generate long reasoning chains. 

\noindent \textbf{Search Space Exploration.} One of the most well-known work is OpenAI's o1 \cite{o1report}, which, although the methodology details remain unpublished, has demonstrated significant improvements in answering mathematical and programming questions by scaling reasoning time and chain length. Several open-source efforts have also explored test-time compute scaling. Repeated sampling methods, such as Best-of-N\cite{brown2024largelanguagemonkeysscaling}, have led to performance gains by repeatedly sampling candidate solutions. Other approaches \cite{jiang2024enhancingllmreasoningrewardguided, qin2024o1replicationjourneystrategic} have investigated constructing reasoning trees to navigate the exploration space of reasoning models and generate higher-quality CoT. Additionally, \cite{xi2024enhancingllmreasoningcritique} have focused on training and utilizing critique models to critically evaluate reasoning results and subsequently generate improved CoT. 

\noindent \textbf{Human-like reasoning strategies} Recent research has introduced self-correction, self-critique, and other human-like mechanisms, allowing one or more models to reflect on and verify existing CoT \cite{xi2024enhancingllmreasoningcritique, liang-etal-2024-encouraging, kamoi-etal-2024-llms}. Moreover, DeepSeek-R1 has leveraged reinforcement learning to incentivize reasoning capability in LLMs, which enables the model to autonomously develop human-like cognitive strategies, such as reflection and backtracking\cite{deepseekai2025deepseekr1incentivizingreasoningcapability}.

\subsection{Knowledge Distillation.}
Utilizing the outputs generated by a more capable model to train a smaller model has become a widely adopted method to improve model performance\cite{hinton2015distillingknowledgeneuralnetwork}. It has been proven that long CoTs distilled from strong reasoning models, can effectively adapt a base model to long CoT reasoning by fine-tuning, significantly enhancing its performance on mathematical problems\cite{qin2024o1replicationjourneystrategic, deepseekai2025deepseekr1incentivizingreasoningcapability, li2025llmseasilylearnreason, ye2025limoreasoning, huang2024o1replicationjourney, zeng2024scalingsearchlearningroadmap}. Furthermore, \cite{ye2025limoreasoning} analyzed the quality of distilled data and demonstrated that a small number of orchestrated demonstrations of cognitive processes can effectively stimulate the model's reasoning capability. Although \cite{deepseekai2025deepseekr1incentivizingreasoningcapability} proposed that conducting reinforcement learning directly from the base model without supervised fine-tuning(SFT) could also elicit long chain reasoning capabilities, incorporating annotated data for cold-start training and SFT further enhances the model's performance and readability of the its responses. Therefore, most recent work follows a paradigm of using annotated data for SFT, followed by reinforcement learning.

\subsection{Token Efficient Chain-of-Thought.}
Although recent works have demonstrated that distillation methods can quickly and cost-effectively enhance the reasoning capabilities of base models, the base model's capability is constrained by teacher models. Meanwhile, the reasoning chains from teacher model are often verbose, repetitive, and may contain logical or computational errors. 

\noindent Previous research has explored ways to think and reason more efficiently. \cite{kimiteam2025kimik15scalingreinforcement} introduced long2shot methods to improve short-CoT models, while \cite{han2024tokenbudgetawarellmreasoning} proposed token-budget-aware reasoning, which dynamically leverages the token budget to guide the reasoning process. Additionally, \cite{li2025llmseasilylearnreason} explored the content accuracy and coherence of reasoning steps, proposing that the structure of long CoTs is key to stimulating a model's reasoning abilities. 

\noindent In this paper, we follow the data distillation and SFT approach, focusing on exploring how to more efficiently leverage distilled long CoTs to activate the model's reasoning capabilities. Specifically, we analyze and deconstruct the patterns of CoTs, identifying and eliminating redundant or erroneous parts while preserving the overall reasoning structure. We also investigate how different components of long CoTs impacts the model’s reasoning performance.

\section{Method}

This section consists of two main parts: a systematic analysis of long CoT distillation methods and the construction of an intelligent optimization framework for long CoT distillation data. We first detail the distillation effectiveness exploration in §~\ref{sec:Distillation}, followed by the introduction of the long CoT distillation data optimization framework DLCoT in §~\ref{sec:Distillation2}.

\subsection{Systematic Analysis of Distillation Methods}
\label{sec:Distillation}
\subsubsection{Data Preparation}
We design a systematic experimental protocol to explore optimal distillation strategies and identify key factors that promote reasoning ability. For the long CoT distillation data, we employee both R1 and QwQ to generate answers for a same set of mathematical problems. To ensure data accuracy, we only retain answers that pass rule-based validation. To ensure data diversity, we incorporate problems from multiple open-source mathematical datasets (see Appendix~\ref{sec:appendix2} Table~\ref{tab:data}). Besides, we utilize Qwen2.5-Math-7B and Llama3.1-8B-Instruct for multiple generations at high temperature, using model pass rates as data difficulty assessment metrics. Finally, we get 33K QwQ generate long CoT data. The R1 distillation dataset comprises 16K self-generated data and 17K data from the open-source Bespoke-Stratos R1 \cite{Bespoke} dataset.
\subsubsection{Distillation Scheme Validation}
The systematic validation we design analyzes from three dimensions: model homology, data scale, and difficulty-diversity. Distillation data from QwQ and R1 were applied to Qwen2.5 series (7B-Math, 14B, 32B) and Llama3.1 series (8B, 70B) models. Detailed experimental results are presented in §~\ref{sec:Experiments1}. 

\subsection{DLCoT}
\label{sec:Distillation2}

\begin{figure*}[!t]
  \centering
  \includegraphics[width=\textwidth, trim=0.5cm 5cm 0.5cm 1cm, clip]{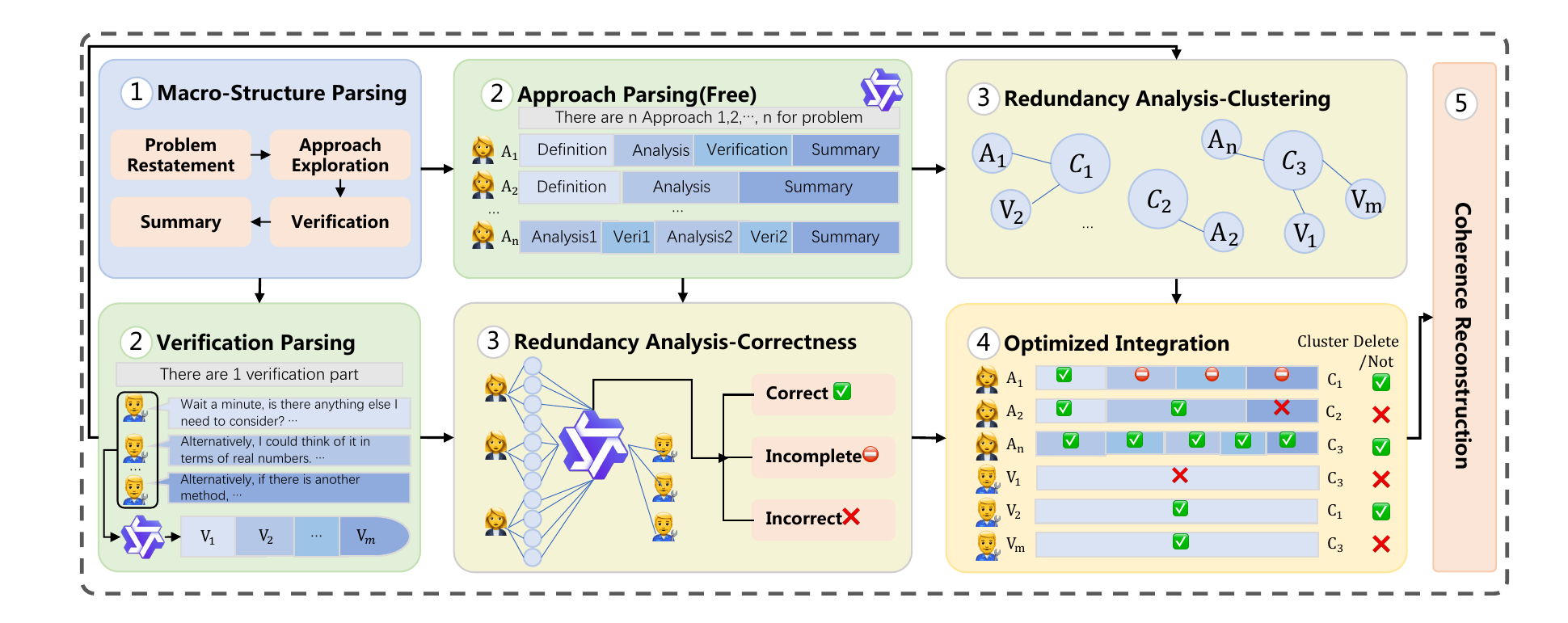}
  \caption{The workflow for DLCoT. It involves five steps: (1) Macro-Structure Parsing,
(2) Approach \& Verification Parsing, (3) Redundancy Analysis, (4) Optimized Integration, and (5) Coherence Reconstruction.}
  \label{fig:DLCoT}
\end{figure*}

\subsubsection{Logic Structure}
To gain deeper insights into the structure of the distilled reasoning chains, we propose Deconstructing Long Chain-of-Thought (DLCoT), a framework that aims to systematically deconstruct long CoT. Our findings reveal that both R1's and QwQ's output share four critical components in their reasoning chains: \textbf{Problem Restatement \& Comprehension} focuses on converting the original problem into a structured statement. \textbf{Approach Exploration} involves searching for potential approaches to solve the problem. This often includes breaking down the problem, formulating problem-solving strategies, and deriving intermediate steps. \textbf{Result Verification} once the answer is reached, this stage involves correctness verification. \textbf{Final Answer} gives a detailed final answer to the problem.
A detailed example can be found in the Appendix~\ref{sec:appendix1}. This structured reasoning framework facilitates clear logical pathways. 
\subsubsection{Approach Structure}
\label{sec:Distillation3}
For the most complex part "approach exploration", we conduct further analysis. The word clouds (Appendix ~\ref{sec:appendixword_cloud} Figure~\ref{fig:word_cloud}) created from this phase suggest that the approach exploration process can be compartmentalized into four distinct stages. The distribution of these stages is distinctly hierarchical, as follows: 
\textbf{Problem Definition} constitutes 3.8\% of the content, primarily represented by Cluster 5. \textbf{Approach Process} dominates the distribution with 84.9\%, identified largely with Cluster 0, which encompasses the computational workflow. \textbf{Verification} accounts for 6.7\% of the distribution, divided into alternative methods (4.9\%, Cluster 1) and stepwise checks (1.8\%, Cluster 2). \textbf{Summary} comprises 4.4\% of the distribution, associated with Cluster 4.

\noindent This hierarchical structure of modules underscores a systematic approach to problem-solving, highlighting the emphasis on computational processes while maintaining critical stages of verification and summary.

\subsubsection{Deconstructing Long Chain-of-Thought}
Building on the above findings, we propose an intelligent Long CoT data optimization framework DLCoT(Figure~\ref{fig:DLCoT}) designed to systematically deconstruct structured reasoning chains, identify critical components, and generate enhanced distillation data. See the Appendix~\ref{sec:appendix4} for more implementation details. The framework comprises five core steps:
\begin{enumerate}
    \item \textbf{Macro-Structure Parsing}: Follows the logic structure analysis result, we first subdivide long CoT data into four parts: Problem Restatement, Approach Exploration, Verification, and Summary.
    \item \textbf{Approach \& Verification Parsing}: LLM-driven autonomous stepwise segmentation under low-constraint conditions. We do not specifically restrict the internal splitting of approaches and reflections, but let the model freely decompose them into structure like §~\ref{sec:Distillation3}.
     \item \textbf{Redundancy Analysis}: We cluster approaches and verifications, based on similarity metrics. Moreover, a three-tier evaluation system was established to assess the quality of these steps: (1) Erroneous Strategies: approaches leading to incorrect final conclusions; (2) Incomplete Strategies: Intermediate derivations with valid steps but fail to reach final conclusions; (3) Correct Strategies: Rigorous derivations yielding accurate and verifiable final answers.
     \item \textbf{Optimized Integration}: We design an experimental protocol (detailed in Section~\ref{sec:Systematic Analysis of DLCoT}) to demonstrate that pruning redundant approaches and eliminating intra-class erroneous strategies significantly enhances distillation efficiency.
     \item \textbf{Coherence Reconstruction}: To ensure fluency and logical consistency in final outputs, we applied minimal-edit rewriting to data modified during redundancy removal. This process preserves structural integrity while eliminating redundancies.
\end{enumerate} 

\section{Experiments}

\begin{table*}[h]
  \centering
  \begin{tabular}{lcccc}
    \hline
    \textbf{Models} & \textbf{AIME2024} & \textbf{MATH500} & \textbf{GSM8K}  & \textbf{OMNIMATH}\\
    \hline
    \verb|QwQ-32B-Preview| & 50.0 & 90.6 & 98.2 & 65.4                          \\
    \verb|Qwen2.5-32B-Instruct| & 20.0 & 82.6 & 95.8 & 39.5                        \\
    \verb|Qwen2.5-7B-Math-QwQ-Distill| & 11.7 & 69.7 & 93.5 &  44.4                        \\
    \verb|Qwen2.5-14B-QwQ-Distill| & 35.0 & 85.3 & 96.7 & 54.2                         \\
    \verb|Qwen2.5-32B-QwQ-Distill| & 46.7 & 91.9 & 97.3 & 60.7                         \\
    \verb|LLaMa3.1-8B-QwQ-Distill| & 8.3 &	47.9  &	86.5 &	23.3    \\\hline
    \verb|DeepSeek-R1| & 79.8 & 97.3 & - & -                         \\
    \verb|Qwen2.5-7B-Math-R1-Distill| & 25.0 & 84.4 & 96.1 & 48.7                        \\
    \verb|Qwen2.5-14B-R1-Distill| & 46.7 & 90.1 & 96.8 & 71.8                         \\
    \verb|Qwen2.5-32B-R1-Distill| & 66.7 & 91.8 & 98.5 &  78.9                         \\
    \verb|Llama-3.1-8B-R1-Distill| & 13.3 & 53.3 & 91.8 & 13.5                         \\
    \hline
  \end{tabular}
  \caption{dataset(1) 6K sample data distillation result
  }
  \label{tab:distill result}
\end{table*}

\subsection{Experimental Setup}
The experimental dataset comprises two components: (1) 16K data samples from the NuminaMath \cite{li2024numinamath}, balanced in terms of source and difficulty level; (2) 17K data samples from the Bespoke-Stratos \cite{bespoke_stratos}. For Dataset(1), we generate answers using QwQ-32B-Preview and Deepseek-R1, with generation parameters in the Appendix~\ref{sec:appendix3}. For Dataset(2), answers are generated by QwQ-32B-Preview using identical parameters, while leveraging open-source Bespoke-Stratos R1 response data. All distillation data pass a rule-based correctness verification.

\noindent Experiments are conducted on the following open-source large language models. Qwen series: Qwen2.5-Math-7B, Qwen2.5-14B, Qwen2.5-32B; Llama series: Llama3.1-8B, Llama3.1-70B. To mitigate random variability, all experiments are repeated three times. Training employs a cosine annealing learning rate scheduler with an initial rate of 1e-5. A uniform batch size of 96 is maintained across experiments to ensure comparative fairness, with training conducted for 6 epochs. For evaluation, models are evaluated by AIME2024, MATH500 \cite{hendrycks2021measuringmathematicalproblemsolving} and GSM8k \cite{cobbe2021trainingverifierssolvemath} benchmarks.

\subsection{Main Results}
\subsubsection{Systematic Analysis of Distillation}
\label{sec:Experiments1}

This study conducts stratified uniform sampling on Dataset (1) across difficulty and source dimensions to obtain 6K data samples for small-batch mathematical knowledge distillation experiments. The experimental results (Table~\ref{tab:distill result}) show that the QwQ dataset exhibits significant advantages on the Qwen series models: the Qwen2.5-32B-QwQ-Distill achieves accuracy rates of 46.67\% and 91.94\% on the AIME2024 and MATH500, respectively. It is observed that Qwen2.5-32B-QwQ-Distill attains comparable mathematical reasoning capabilities to QwQ-32B-Preview, and significantly outperforms the baseline Qwen2.5-32B-Instruct. However, when applied to the Llama series, the model performance significantly deteriorates across all benchmarks except GSM8K. In Contrast, the R1 distillation dataset exhibits superior adaptability. In R1 distillation experiments with Qwen2.5-14B, the performance of the 14B model exceeds that of the QwQ-32B-Preview.

\subsubsection{Systematic Analysis of DLCoT}
\label{sec:Systematic Analysis of DLCoT}

\begin{table*}[h]
    \centering
    \caption{Reduce Redundancy Ablation Study Result}
    \begin{tabular}{@{} ll|cc|cc|cr  @{}} 
        \toprule
        \textbf{Models} && \multicolumn{2}{c|}{\textbf{AIME2024}} & \multicolumn{2}{c|}{\textbf{MATH500}} & \multicolumn{2}{c}{\textbf{GSM8K}} \\
        \cmidrule(lr){3-4} \cmidrule(lr){5-6}\cmidrule(lr){7-8}
         & & Accuracy & Token & Accuracy & Token & Accuracy & Token \\
        \midrule
        \multicolumn{8}{c}{\textbf{\textit{QwQ Distilled LLMs}}} \\ 
        \midrule
                 &Qwen2.5-14B&  &  & 80 &  & 94.8 &   \\
                 &  +baseline& 33.3 & 15095.2 & 85.3 & 3702.5 & 96.2 &  1071.3 \\
                 &  +DLCoT-multi1& 33.3 & 18833.3 & 85.3 & 4031.3 & 96.7 &  1109.1  \\
                 &  +DLCoT-multi2& 35.0 & 14881.8 & 84.6 & 3899.7 & 96.4 & 1023.6  \\
                 &  +DLCoT-multiall& \textbf{40.0} & 14699.1 & \textbf{87.7} & 3547.4 & \textbf{96.8} &  1049.3 \\
        \cmidrule(lr){3-8}
        Reduce Redundancy &Llama3.1-70B&  &  & 68.0 &  & 95.1 &   \\
                 &  +baseline        & 16.7 & 24799.8 & 65.3 & 12697.1  & 93.6 & 3069.9  \\
                 &  +DLCoT-multi1& 16.7 & 20528.5 & 65.3 & 9195.0 & 92.7 &  1867.2 \\
                 &  +DLCoT-multi2& \textbf{23.3} & 22127.1 & 68.3 & 8271.2 & 93.2 & 937.54  \\
                 &  +DLCoT-multiall& \textbf{23.3} & 22674.1 & \textbf{69.7} & 2889.1 & \textbf{95.9} & 1007.0  \\
        
        \midrule
        
        \multicolumn{8}{c}{\textbf{\textit{R1 Distilled LLMs}}} \\ 
        \midrule
                 &Qwen2.5-14B&  &  & 80 &  & 94.8 &   \\
                 &  +baseline& 46.7 & 28526.5 & \textbf{91.7} & 5244.2 & \textbf{97.6} & 1457.0  \\
                 &  +DLCoT-multi1& 38.3 & 19690.1 & 88.5 & 5257.3 & 96.3 &  1422.0 \\
                 &  +DLCoT-multi2& 38.3 & 19827.7 & 90.1 &  5035.0 & 96.9 &  1425.7 \\
                 &  +DLCoT-multiall& \textbf{53.3} & 18825.0 & 91.4 & 4978.3 & 96.8 &  1238.6 \\
        \cmidrule(lr){3-8}
        Reduce Redundancy &Llama3.1-8B&  &  & 51.9 &  & 84.5 &   \\
                 &  +baseline        & 6.7 & 29229.8 & 61.7 & 25394.6 & 93.0 &  16610.5 \\
                 &  +DLCoT-multi1& 6.7 & 29231.1 & 61.1 & 18503.5 & 92.8 &  5120.3 \\
                 &  +DLCoT-multi2& 5.0 & 29143.5 & 61.8 & 16705.3 & 90.1 &  3799.0 \\
                 &  +DLCoT-multiall& \textbf{8.3} & 29191.5 & \textbf{62.8} & 14763.5 & \textbf{93.6} & 3795.6  \\
        \bottomrule
    \end{tabular}
    \label{tab:ablation}
\end{table*}

Based on the DLCoT methodology, we examine the most effective strategies for Step 4, "Optimized Integration", to address the question: "What are the critical components of long CoT methods?" We propose that long CoT includes a logical core and suggest that removing redundant, erroneous, or ineffective paths can improve both distillation efficacy and token efficiency.

\begin{table*}[h]
    \centering
    \caption{Reduce Incorrectness Ablation Study Result}
    \begin{tabular}{l|c|c|c} 
        \toprule
        \textbf{Models} & \multicolumn{1}{c|}{\textbf{AIME2024}} & \multicolumn{1}{c|}{\textbf{MATH500}} & \multicolumn{1}{c}{\textbf{GSM8K}} \\
        \midrule
        \multicolumn{4}{c}{\textbf{\textit{QwQ Distilled LLMs}}} \\ 
        \midrule
                 Qwen2.5-14B+DLCoT-multiall & 40 & 87.7 & 96.8 \\
                 Qwen2.5-14B+DLCoT-multiall-incorrectness & 33.3 & 84.5 & 96.3  \\
                 Llama3.1-70B+DLCoT-multiall & 23.3 & 69.7 & 95.9  \\
                 Llama3.1-70B+DLCoT--multiall-incorrectness & 15.0 & 62.8 & 95.5 \\
        \midrule        
        \multicolumn{4}{c}{\textbf{\textit{R1 Distilled LLMs}}} \\ 
       \midrule
             Qwen2.5-14B+DLCoT-multiall& 53.3 & 91.4 & 96.8 \\
             Qwen2.5-14B+DLCoT-multiall-incorrectness & 40.0 & 88.2 & 93.8 \\
             Llama3.1-8B+DLCoT-multiall & 8.3 & 62.8 & 93.6 \\
             Llama3.1-8B+DLCoT-multiall-incorrectness & 6.7 & 65.5  & 94.8  \\
        \bottomrule
    \end{tabular}
    \label{tab:incorrectness}
\end{table*}

\noindent\textbf{Redundancy Reduction.} Through redundancy analysis, we identify substantial repetitive approaches within answers. Our ablation study is designed to progressively remove 1, 2, or all redundant approaches, denoted as DLCoT-multi1, DLCoT-multi2 and DLCoT-multiall), in reverse order while ensuring approach diversity by retaining at least one approach per cluster. Experimental results demonstrate that maximal reduction of redundant approaches is the best for both R1 and QwQ datasets, Table~\ref{tab:ablation}.

\noindent\textbf{Incorrectness Reduction.} Building on the removal of all redundant methods, we aim to explore further optimizations. In long CoT, we observed numerous computational and derivation errors. A straightforward approach is to delete methods that are marked as erroneous, denoted as DLCoT-multiall-incorrectness, while ensuring that the trajectory leading to the correct answer is preserved.
The experimental results are shown in Table ~\ref{tab:incorrectness}. As observed, further removing erroneous approaches and incorrect steps, in addition to eliminating redundant methods, have a slight negative impact on the model performance. Moreover, this negative effect increased with difficulty of problems, AIME > MATH > GSM8K.

\subsection{Analysis}

\begin{figure}[h]
  \centering
  \includegraphics[width=\columnwidth]{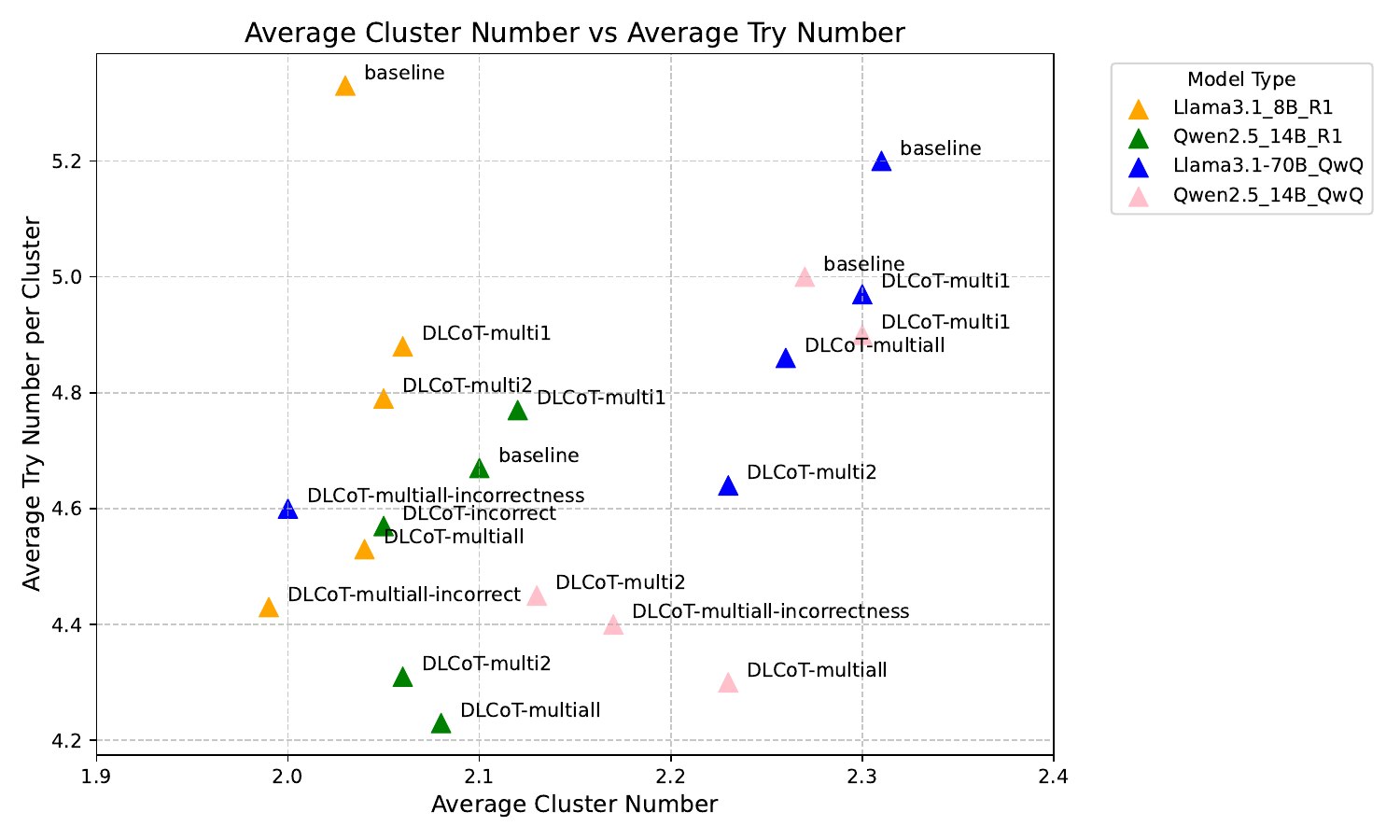}
  \caption{Average Cluster Number v.s. Average Try Number per Cluster.}
  \label{fig:cluster_try}
\end{figure}

\textbf{Approach Diversity.} Try count is the sum of approach count and verification count. Figure~\ref{fig:cluster_try} reflect the relationship between the average cluster count and the average try count per cluster for each distillation model on the AIME, MATH, and GSM8K datasets. The slope (=Try count/cluster count) can reflect the diversity of answers, the smaller the slope, the greater the diversity. It can be seen that the DLCoT-multiall method shows the lowest slope in our ablation study. At the same time, DLCoT-multiall-incorrectness is the one that significantly reduces the average number of clusters. Removing erroneous methods reduces the diversity of approaches in the training dataset, which negatively impact the approach exploration in generation. This shows that the non-redundant reasoning that maintains the diversity of the trunk solution in long CoT is the core of its ability to stimulate reasoning.

\begin{figure}[h]
  \centering
  \includegraphics[width=\columnwidth]{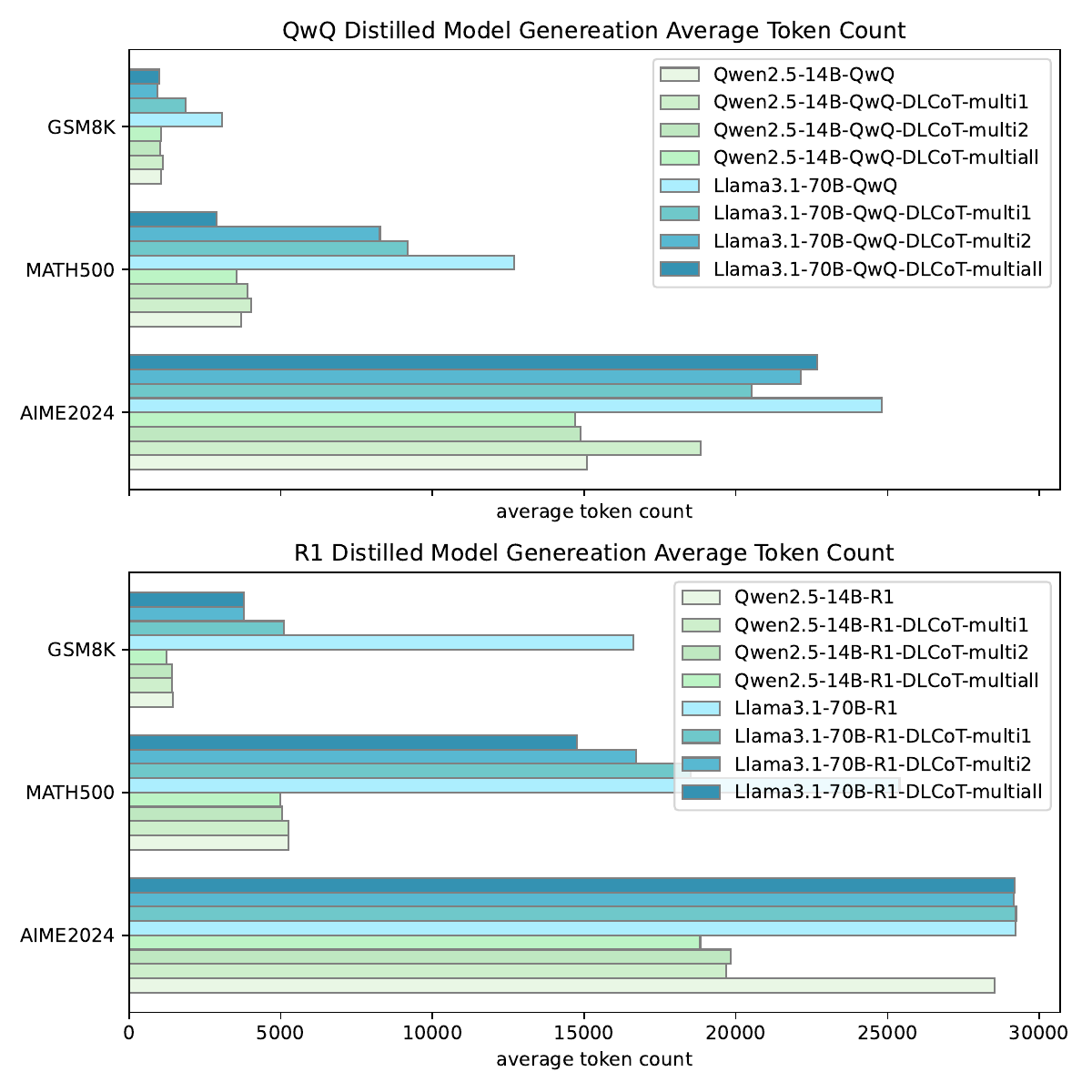}
  \caption{Average token output of various distillation models on AIME2024, MATH500 and GSM8K.}
  \label{fig:Average_Token}
\end{figure}

\noindent\textbf{Token Efficiency.} Figure~\ref{fig:Average_Token} shows that the number of tokens in DLCoT-multiall is significantly smaller than the original version, which directly shows that DLCoT improves output's token efficiency. However, if only some duplicate solutions are deleted, the same effect may not be achieved. We also observe that tokens increases with the difficulty of the problem (AIME> MAH> GSM8K). This shows that there is a direct correlation between problem complexity and resource consumption. Regarding the comparison between models, the number of tokens generated by Llama is significantly larger than that of Qwen. 

\noindent\textbf{Incorrectness Failure.} Deleting steps incorrect calculation and derivation errors can affect the logical coherence of CoTs and disrupt important existing cognitive strategies. Besides, We observed that the error deletion strategy disrupted many mistakes that were later corrected, thereby undermining the integrity of the model's self-reflection process. This finding aligns with the conclusions drawn in \cite{li2025llmseasilylearnreason}, where the compromised structure of long CoT significantly negatively impacted model's training to elicit strong reasoning capability.

\section{Conclusion}
In this paper, we systematically investigate the pattern and optimization strategies for long CoT distillation in large language models (LLMs). We propose DLCoT, an framework designed to optimize distillation data through intelligent segmentation, redundancy elimination, and error correction. Through extensive experimentation and analysis, we uncover key insights into the structure of long CoTs as well as the factors influencing eliciting reasoning capabilities. Our approach achieves significant improvements and token reduction in accuracy across multiple benchmarks, demonstrating its practical utility in real-world applications.

\section{Limitations}
While this study provides valuable insights into the optimization of long chain-of-thought (CoT) distillation, there are several areas that remain unexplored:

\noindent\textbf{Generalizability of Distillation Data Across Models.} We haven't thoroughly investigate the transferability of reasoning capabilities through distilled data when applied to models with significantly different design paradigms or training objectives. A deeper understanding of how distilled data interacts with heterogeneous architectures can further refine cross-model knowledge transfer.

\noindent\textbf{Impact on Reinforcement Learning Tasks.} The potential of DLCoT-processed data to accelerate improvements in reinforcement learning (RL) tasks has not been examined. It remains an open question whether models trained on DLCoT-optimized data can achieve faster convergence or superior performance when subjected to RL-based training.

\bibliography{references}

\appendix

\section{Segment Example}
\label{sec:appendix1}

\begin{lstlisting}[language=json, caption={Example JSON}]
{
"Question_Repeat": "So I need to prove that the line containing the midsegment of a triangle is parallel to one of its sides and that the midsegment is half as long as that side. Okay, first things first, I need to make sure I understand what a midsegment is.\n\nA midsegment of a triangle is a segment that connects the midpoints of two sides of the triangle. So, if I have a triangle, say, ABC, and I pick two sides, AB and AC, and find their midpoints, let's call them D and E respectively, then the midsegment would be DE.\n\nAlright, so DE connects the midpoints D and E of sides AB and AC. Now, I need to show two things:\n\n1. The line containing DE is parallel to the third side, which is BC.\n\n2. The length of DE is half the length of BC.\n\nOkay, let's tackle the first part: proving that DE is parallel to BC.", 
"Problem_Understand": "\n\nI remember that in triangles, if a line is parallel to one of the sides, it divides the other two sides proportionally. That sounds like the Triangle Proportionality Theorem. Maybe I can use that here.\n\nThe Triangle Proportionality Theorem states that if a line is parallel to one side of a triangle and intersects the other two sides, then it divides those sides proportionally.\n\nBut wait, in this case, I have the midsegment DE, which connects the midpoints D and E of AB and AC. So, AD = DB and AE = EC, meaning AD/AB = 1/2 and AE/AC = 1/2.\n\nSo, in this case, DE divides AB and AC in the same ratio, which is 1:1 on each side.\n\nHmm, actually, since D and E are midpoints, AD = DB and AE = EC, so AD/AB = 1/2 and AE/AC = 1/2.\n\nWait a minute, the Triangle Proportionality Theorem says that if a line is parallel to one side and intersects the other two, it divides them proportionally.\n\nIn this case, DE divides AB and AC each in the ratio 1:1, but AB and AC are not the same length necessarily, so I'm a bit confused.\n\nAlternatively, maybe I can use similar triangles to prove this.\n\nLet me try that approach.\n\nIf I can show that triangle ADE is similar to triangle ABC, then corresponding sides are proportional, and corresponding angles are equal, which would help me establish parallelism.\n\nSo, let's see: in triangle ADE and triangle ABC,\n\n- Angle A is common to both triangles.\n\n- If I can show that angle ADE is equal to angle ABC, and angle AED is equal to angle ACB, then the triangles are similar by AA similarity.\n\nBut how do I show that the angles are equal? Maybe I need to use some properties of midsegments or perhaps consider vectors.\n\nAlternatively, maybe I should use coordinate geometry to prove this. I can assign coordinates to the vertices of the triangle and then calculate the slopes and lengths to show the required properties.", 
"Solution_Explore": {
    "Solution1": {
        "Coordinate Geometry Approach": "\n\nLet's try that.\n\nLet me place triangle ABC on a coordinate plane.\n\nLet's assume point A is at \(0,0), point B is at (2b,0), and point C is at (2c,2h), where b, c, and h are real numbers.\n\nSo, coordinates:\n\n- A: (0,0)\n\n- B: (2b,0)\n\n- C: (2c,2h)\n\nNow, find the midpoints D and E of AB and AC, respectively.\n\nMidpoint D of AB: ((0 + 2b)/2, (0 + 0)/2) = (b, 0)\n\nMidpoint E of AC: ((0 + 2c)/2, (0 + 2h)/2) = (c, h)\n\nNow, the midsegment DE connects D(b,0) to E(c,h).\n\nFirst, find the slope of DE.\n\nSlope of DE: (h - 0)/(c - b) = h/(c - b)\n\nNow, find the slope of BC.\n\nCoordinates of B: (2b,0), coordinates of C: (2c,2h)\n\nSlope of BC: (2h - 0)/(2c - 2b) = 2h/(2c - 2b) = h/(c - b)\n\nHey, the slopes are equal! That means DE is parallel to BC.\n\nSo, part one is proved: the midsegment DE is parallel to side BC.\n\nNow, for the second part: proving that DE is half the length of BC.\n\nLet's calculate the lengths.\n\nFirst, length of BC.\n\nCoordinates of B: (2b,0), coordinates of C: (2c,2h)\n\nDistance formula: sqrt((2c - 2b)^2 + (2h - 0)^2) = sqrt(4(c - b)^2 + 4h^2) = sqrt(4[(c - b)^2 + h^2]) = 2 * sqrt((c - b)^2 + h^2)\n\nNow, length of DE.\n\nCoordinates of D: (b,0), coordinates of E: (c,h)\n\nDistance formula: sqrt((c - b)^2 + (h - 0)^2) = sqrt((c - b)^2 + h^2)\n\nComparing the two lengths:\n\nLength of DE: sqrt((c - b)^2 + h^2)\n\nLength of BC: 2 * sqrt((c - b)^2 + h^2)\n\nTherefore, DE = (1/2) * BC.\n\nSo, the midsegment DE is half the length of BC.\n\nAlright, that was straightforward using coordinate geometry."}, 
    "Solution2": {
        "Similar Triangles Approach": "\n\nBut maybe I can also prove this using similar triangles without coordinates.\n\nLet me try that approach.\n\nConsider triangle ABC again.\n\nLet D be the midpoint of AB, and E be the midpoint of AC.\n\nDraw DE.\n\nI need to show that DE is parallel to BC and DE = (1/2)BC.\n\nLet me consider triangles ADE and ABC.\n\nWe have:\n\n- AD = DB, so AD = (1/2)AB\n\n- AE = EC, so AE = (1/2)AC\n\nNow, in triangles ADE and ABC,\n\n- AD/AB = 1/2\n\n- AE/AC = 1/2\n\n- Angle A is common to both triangles.\n\nWait, does this imply that triangles ADE and ABC are similar with a ratio of 1:2?\n\nIf so, then corresponding sides would be in the ratio 1:2, which would mean DE/BC = 1/2, so DE = (1/2)BC.\n\nAlso, corresponding angles would be equal, which would imply that DE is parallel to BC.\n\nBut is this enough to establish similarity?\n\nI think I need to confirm the similarity.\n\nAlternatively, maybe I can use the Basic Proportionality Theorem (also known as Thales' theorem), which states that if a line is drawn parallel to one side of a triangle to intersect the other two sides in distinct points, the other two sides are divided in the same ratio.\n\nBut in this case, I have the ratios AD/AB = AE/AC = 1/2, and I need to show that DE is parallel to BC.\n\nWait, is there a way to use the converse of the Basic Proportionality Theorem here?\n\nThe converse states that if a line divides two sides of a triangle in the same ratio, then it is parallel to the third side.\n\nIn this case, AD/AB = AE/AC = 1/2, so by the converse of the Basic Proportionality Theorem, DE is parallel to BC.\n\nThat seems straightforward.\n\nSo, DE is parallel to BC.\n\nNow, for the lengths.\n\nSince DE is parallel to BC and divides AB and AC in the ratio 1:1, then by the properties of similar triangles, the ratio of similarity is 1:2, so DE = (1/2)BC.\n\nAlternatively, since DE is parallel to BC, angle ADE = angle ABC and angle AED = angle ACB (corresponding angles), so triangles ADE and ABC are similar by AA similarity.\n\nTherefore, corresponding sides are in proportion, so DE/BC = AD/AB = AE/AC = 1/2.\n\nThus, DE = (1/2)BC.", 
        "Slope Calculation and Parallelism": "\n\nSo, both parts are proved using similar triangles without coordinates.\n\nI think this is a solid proof.\n\nLet me just double-check.\n\nGiven: D and E are midpoints of AB and AC, respectively.\n\nTo prove:\n\n1. DE is parallel to BC.\n\n2. DE = (1/2)BC.\n\nProof:\n\nIn triangle ABC, let D and E be the midpoints of AB and AC, respectively.\n\nThen, AD = DB and AE = EC, so AD = (1/2)AB and AE = (1/2)AC.\n\nBy the converse of the Basic Proportionality Theorem, since AD/AB = AE/AC = 1/2, it follows that DE is parallel to BC.", 
        "Length Calculation and Proportionality": "\n\nNow, since DE is parallel to BC and triangles ADE and ABC have corresponding angles equal (angle A is common, angle ADE = angle ABC, and angle AED = angle ACB), they are similar by AA similarity.\n\nTherefore, the ratio of corresponding sides is equal: DE/BC = AD/AB = 1/2.\n\nThus, DE = (1/2)BC.\n\nQ.E.D.\n\nYes, that seems correct."
        }
    }, 
    "Verify": {
        "Self-Affirmation": "\n\nAlternatively, using vector geometry could also prove this, but coordinate geometry was simpler in this case.\n\nI think I've covered both parts of the proof adequately."}, 
    "Conclusion": "\n\n**Final Answer**\n\n\\[ \\boxed{DE \\parallel BC \\text{ and } DE = \\frac{1}{2}BC} \\]."
    }
}
\end{lstlisting}

\section{DLCoT Implementation Instructions}
\label{sec:appendix4}

\begin{tcolorbox}[colback=white!95!gray,colframe=gray!50!black,rounded corners, label={Macro-Structure Parsing}, title={Macro-Structure Parsing}] 
You are a mathematical solution structure decomposition agent. Your task is to analyze a mathematical problem's solution, then restructure it into a specific format following these rules:

1. Split the solution into exactly 5 sequential components:
   
   - Question\_Repeat: The initial statement of the problem, including the "let's break this down step by step" part
   
   - Problem\_Understand: Only the initial high-level analysis before diving into calculations (if present, otherwise skip)
   
   - Approach\_Explore: The main solution process, including all calculations and intermediate steps up to finding the first answer
   
   - Verify: Include ALL verification steps, alternative approaches, and checking calculations after the initial approach (if present, otherwise skip)
   
   - Conclusion: Include both the final concluding statement AND the boxed answer

2. Natural Transition Points:

   - Question\_Repeat → Problem\_Understand: Break after the problem is stated and before analysis begins
   
   - Problem\_Understand → Approach\_Explore: Break after conceptual analysis and before first calculations
   
   - Approach\_Explore → Verify: Break after obtaining first answer and before starting verification
   
   - Verify → Conclusion: Break after all checking is complete and before final statement

3. Format Requirements:

   - Present the output in two main sections: "\# Answer Split" and "\# Structure"
   
   - Under "Answer Split", use "\#\#" headings for each component (Question\_Repeat, Problem\_Understand, etc.)
   
   - Include the exact original text under each heading, preserving all line breaks and formatting
\end{tcolorbox}   
\begin{tcolorbox}
[colback=white!95!gray,colframe=gray!50!black,rounded corners, label={Macro-Structure Parsing}, title={Macro-Structure Parsing}]   
   - After all components, add the "\# Structure" section with the array of component names
   
   - Ensure no text is truncated or modified from the original

4. Content Distribution Guidelines:

   - Question\_Repeat must include both the problem statement AND any initial "let's break this down" statement
   
   - Problem\_Understand should be limited to only the initial analysis before any calculations
   
   - Approach\_Explore should contain all mathematical steps, intermediate checking and calculations
   
   - Place ALL verification steps, alternative approach, and checking calculations in the Verify section
   
   - Conclusion should contain only the final answer with proper \\boxed{} notation

5. Critical Requirements:

   - Preserve all original mathematical notation exactly, especially \\boxed{} notation
   
   - Maintain all line breaks as they appear in the original text
   
   - Include all text exactly as written without any modifications
   
   - Ensure each section break occurs at natural transition points in the approach
   
   - Ensure all verification steps are in the Verify section

\# Input
   
\#\# 1. The mathematics question's solution

[solution input here]

\#\# Output
\end{tcolorbox}

\begin{tcolorbox}[colback=white!95!gray,colframe=gray!50!black,rounded corners, label={Approach Parsing}, title={Approach Parsing}] 
As an AI assistant, your task is to restructure mathematical solution text into a hierarchical format. Follow these steps:

1. Parse the input text and organize it into the following structure:

   - Top level: "Approach Explore Split" (main heading)
   
   - Second level: Approach (\#\# Approach1, \#\# Approach2, etc.)
   
   - Third level: Analysis components

2. Format Rules:

   - Use \# for main heading
   
   - Use \#\# for Approach level
   
   - Use \#\#\# for component headers
   
   - DO NOT use original text content as component headers
   
   - Preserve all mathematical notations and equations
   
   - Maintain original text content within appropriate sections

3. After the main content, add a "structure" section that summarizes the hierarchy using: Approach[n]: [list of components]

4. Approach Separation Rules:

    - Start a new Approach section when a different approach to the same problem is attempted
    
    - The strategy fundamentally changes
    
    - Keywords: "alternatively", "Maybe there's a better way."

5. Content Preservation:

    - Keep all mathematical notations (\( \) and LaTeX)
    
    - Use exact text as it appears
    
    - Maintain all numerical values and equations
    
    - Keep logical flow intact
    
    - Include all text exactly as written without any modifications

Please format the following mathematical solution accordingly :

\# Mathematical Approach:

[solution input here]

Remember double check original mathematical solution hasn't been rewritten.
\end{tcolorbox}

\begin{tcolorbox}[colback=white!95!gray,colframe=gray!50!black,rounded corners, label={Approach Parsing}, title={Verification Parsing}] 
Given a mathematical solution and its reflection text, identify and categorize the verification steps into specific categories. The output should contain two parts:

1. A formatted section titled "\# Verify Split" containing:

   - Each verification step as a second-level heading (\#\#)
   
   - The relevant text under each category keeping the mathematical notation intact
   
   - Separate the content of the self-talk affirmation and negation programs into the self-affirmation/self-negation
   
   - "self-affirmation" example: I think this is solid/ Yes, that checks out
   
   - "self-negation" example: We already did that / but that might be too complicated

2. A section titled "\# structure" containing:

   - A simple list of the verification categories in the exact order they appear in the text
   
   - Format: ["category1", "category2", ...]

Key Guidelines:

- Include complete verification sequences even when they span multiple paragraphs

- Keep all mathematical notation and calculations exactly as they appear

- Maintain the logical flow of verification steps

- Focus on numerical verification and constraint checking

- Include all text exactly as written without any modifications

- Include complete verification sequences even when they span multiple paragraphs

Format the output exactly as shown:
\# Verify Split

\#\# [Category\_Name]

[Complete verification text with all calculations]

\#\# [Next\_Category\_Name]

[Complete verification text with all calculations]

\# structure

["category1", "category2", ...]

\end{tcolorbox}
\begin{tcolorbox}
[colback=white!95!gray,colframe=gray!50!black,rounded corners, label={Verification Parsing}, title={Verification Parsing}]

Important: Only INCLUDE the mathematical REFLECTION TEXT, NOT the SOLUTION TEXT itself.

\#\# Input

\#\#\# 1. Solution Text:

[solution input here]

\#\#\# 2. Please split the following mathematical reflection text:

[reflection input here]

\#\#Output: 
\end{tcolorbox}

\begin{tcolorbox}[colback=white!95!gray,colframe=gray!50!black,rounded corners, label={Redundancy Analysis}, title={Redundancy Analysis}] 
You are a professional mathematics teacher tasked with evaluating student solutions to mathematical problems. I will provide you with:

1. A mathematical problem

2. The standard solution for this problem

3. Multiple solutions that need evaluation

For each solution, you need to carefully analyze and provide two labels:

\# Label 1: Evaluate Completeness and Correctness

Analyze whether each solution fully derives the final answer to the question and whether the final answer matches the final answer marked with \\boxed in the standard solution. 
Note that in label 1, we only care about whether the final answer in solution matches the final answer marked with \\boxed in the standard solution. There could be errors in the solution, but as long as the final answer matches, it is considered correct.

- If the solution fully derives the final answer to the question, and matches the final answer marked with \\boxed in the standard solution, output: <label1>Correct</label1>

- If the solution fully derives a final answer to the question, but differs from the final answer marked with \\boxed in the standard solution, output: <label1>Incorrect</label1>

- If the solution is not complete and does not fully derive the final answer to the question, output: <label1>Incomplete</label1>
\end{tcolorbox}

\begin{tcolorbox}
[colback=white!95!gray,colframe=gray!50!black,rounded corners, label={Redundancy Analysis}, title={Redundancy Analysis}]
- Note that the format of the final answer in the solution may have slightly different representations compared to the final answer in the standard solution. For numerical or formula solutions, if they are mathematically equivalent, they are considered correct. For example, 109.2 and \\frac{{546}}{{5}} are equivalent and thus correct.

\# Label 2: Evaluate Calculation and Derivation Errors

Even though the solution may be correct, incorrect, or incomplete as defined above, there might still be Calculation and Derivation Errors in its derivation process.

- If there are calculation and derivation errors, output: <label2>Calculation and Derivation Error</label2>, 

- Then in the next line, talk about the explanation for the Calculation and Derivation Error.

- Then in the next line, quote the erroneous parts from the solution completely and exactly without omitting any words. An erroneous part could be a step or several steps. You should fully include where the error starts and ends.

- Note that the part you quote must exactly match a portion of the solution. Do not add any extra characters, including newline characters, spaces, etc.

- if the the solution does not contain any calculation and derivation errors, output: <label2>No Calculation and Derivation Error</label2>

\# Output Format

For each solution, provide output in the following format:

\#\# Solution X (where X is the solution number)

[Label 1]

Explanation for label1: [Detailed explanation of the reason for Label 1]

[Label 2]

Explanation for label2: [Detailed explanation of the reason for Label 2]

Quoted erroneous parts: [Quoted erroneous parts from the solution]

\# Evaluation Principles

1. Examine each step of every solution carefully
\end{tcolorbox}

\begin{tcolorbox}
[colback=white!95!gray,colframe=gray!50!black,rounded corners, label={Redundancy Analysis}, title={Redundancy Analysis}]

2. Provide specific and clear explanations, avoiding vague statements

3. Note when evaluating solutions, treat each solution as a complete independent answer. Do not make connections between multiple solutions.

4. You must strictly follow the format of the output

Remember to maintain consistency in your evaluation across all solutions while being thorough in your analysis of each specific case.

\# Input

Question:

[question input here]

Standard solution:

[standard solution input here]

Solutions to be evaluated:

[solution input here]

\# Output: 
\end{tcolorbox}

\begin{tcolorbox}[colback=white!95!gray,colframe=gray!50!black,rounded corners, label={Optimized Integration}, title={Optimized Integration}] 
Given a mathematical solution and its reflection text, identify and categorize the verification steps into specific categories. The output should contain two parts:

1. A formatted section titled "\# Verify Split" containing:

   - Each verification step as a second-level heading (\#\#)
   
   - The relevant text under each category keeping the mathematical notation intact
   
   - Separate the content of the self-talk affirmation and negation programs into the self-affirmation/self-negation
   
   - "self-affirmation" example: I think this is solid/ Yes, that checks out
   
   - "self-negation" example: We already did that / but that might be too complicated

2. A section titled "\# structure" containing:

   - A simple list of the verification categories in the exact order they appear in the text
   
   - Format: ["category1", "category2", ...]
\end{tcolorbox}

\begin{tcolorbox}
[colback=white!95!gray,colframe=gray!50!black,rounded corners, label={Optimized Integration}, title={Optimized Integration}]
Key Guidelines:

- Include complete verification sequences even when they span multiple paragraphs

- Keep all mathematical notation and calculations exactly as they appear

- Maintain the logical flow of verification steps

- Focus on numerical verification and constraint checking

- Include all text exactly as written without any modifications

- Include complete verification sequences even when they span multiple paragraphs

Format the output exactly as shown:

\# Verify Split

\#\# [Category\_Name]

[Complete verification text with all calculations]

\#\# [Next\_Category\_Name]

[Complete verification text with all calculations]

\# structure

["category1", "category2", ...]

Important: Only INCLUDE the mathematical REFLECTION TEXT, NOT the SOLUTION TEXT itself.

\#\# Input

\#\#\# 1. Solution Text:

[solution input here]

\#\#\# 2. Please split the following mathematical reflection text:

[reflection input here]

\#\#Output: 
\end{tcolorbox}

\section{Distillation Data Generation Parameters}
\label{sec:appendix3}
We use the QwQ-32B-Preview and DeepSeek-R1 models to generate distillation data at 33K and 16K, respectively. The model parameters are listed in the following Table~\ref{tab:data}. For QwQ, we apply offical system prompt: "You are a helpful and harmless assistant. You are Qwen developed by Alibaba. You should think step-by-step.". For R1 we don't use any system prompt.
\begin{table}[t]
  \centering
  \begin{tabular}{lcc}
    \hline
    \textbf{Parameters} & \textbf{QwQ-32B-Preview} & \textbf{DeepSeek-R1} \\
    \hline
    Top-k & 1 & 1 \\
    Top-p & 0.5 & 0.5 \\
    Temperature & 0.5 & 0.5 \\
    Max new tokens & 32,768 & 32,768 \\
    \hline
  \end{tabular}
  \caption{Long CoT Distillation Data Generation Parameters}
  \label{tab:data}
\end{table}

\section{Dataset Distribution}
\label{sec:appendix2}
We provide the sources of the prompts that used in our experiments. The prompts consist of two parts, first part is 16K NuminaMath prompts. For GSM8k \cite{cobbe2021trainingverifierssolvemath}, MATH \cite{hendrycks2021measuringmathematicalproblemsolving} and ORCAMATH\cite{mitra2024orcamathunlockingpotentialslms} we get prompts from their original paper. The other dataset is 17K data from the open-source Bespoke-Stratos R1 (Be-264
spoke, 2025) dataset.
\begin{table}[h]
  \centering
  \begin{tabular}{lc}
    \hline
    \textbf{Source} & \textbf{Data Count} \\
    \hline
    \verb|NuminaMath cn_k12|    & 1772           \\
    \verb|gsm8k|    & 2852           \\
    \verb|math|    & 3000           \\
    \verb|orca_math|    & 1814           \\
    \verb|NuminaMath amc_aime|    & 574           \\
    \verb|NuminaMath aops_forum|    & 1552           \\
    \verb|NuminaMath olympiads|    & 3000           \\
    \verb|NuminaMath synthetic_math|     & 176            \\
    \verb|NuminaMath synthetic_math|    & 1314           \\
    \verb|Bespoke-Stratos|    & 16710           \\\hline
  \end{tabular}
  \caption{Long CoT Data Composition}
  \label{tab:data}
\end{table}



\section{Approach Word Clouds}
\label{sec:appendixword_cloud}
\begin{figure}[h]
  \centering
  \includegraphics[width=\columnwidth, trim=0cm 4.5cm 0cm 4cm, clip]{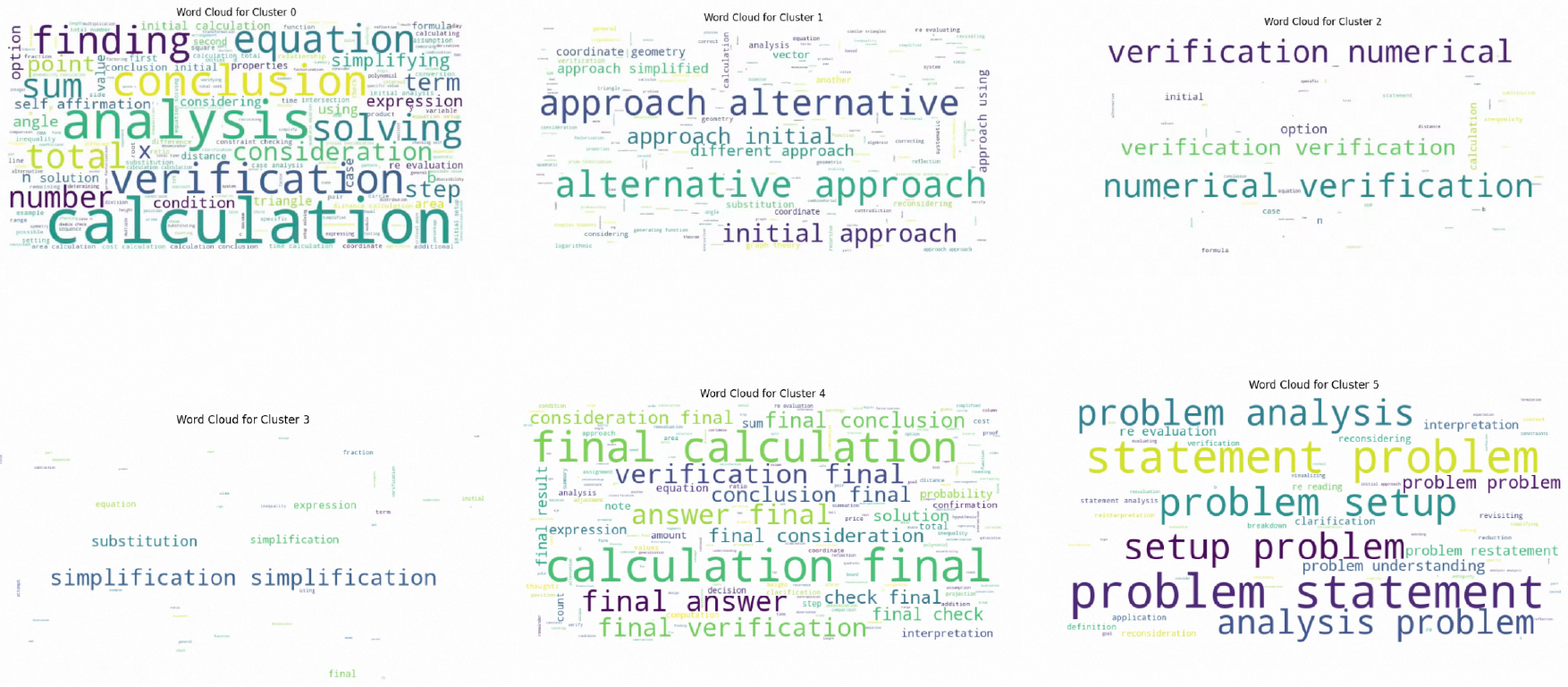}
  \caption{Word Clouds of Approach Exploration.}
  \label{fig:word_cloud}
\end{figure}

\end{document}